\documentclass[letterpaper, 10 pt, conference]{ieeeconf}  %

\IEEEoverridecommandlockouts                              %

\usepackage{amsmath} %
\usepackage{amssymb}  %

\usepackage{tikz} %
\usepackage{pgfplots}
\pgfplotsset{compat=1.18}
\usepackage{bm} %
\usepackage{pxfonts} %
\usepackage{multirow}
\usepackage{tabularx}
\usepackage{subcaption}
\usepackage{math_symbols}

\definecolor{green}{RGB}{0, 0, 0} %
\definecolor{greenCmd}{RGB}{0, 0, 0}
\definecolor{blueEst}{RGB}{0, 0, 0}
\definecolor{graphBlue}{HTML}{6C8EBF}
\definecolor{turquoise}{RGB}{0, 0, 0}
\definecolor{teal}{HTML}{00796b} %

\newcommand{\greenCmd}[1]{{\color{greenCmd}{#1}}}

\newcommand{\blueEst}[1]{{\color{blueEst}{#1}}}

\title{\LARGE \bf
Enabling Scalable Kinesthetic Teaching via Observer-based Hand-guiding with Active Support
}

\author{Anna Tuma$^{1,2}$, Giuseppe Monetti$^{1}$, Jochen J. Steil$^{2}$ and Niels Dehio$^{1,2}$%
\thanks{Accepted for publication at the 2026 IEEE International Conference on Robot and Human Interactive Communication (RO-MAN). \copyright~2026 IEEE. Personal use of this material is permitted. Permission from IEEE must be obtained for all other uses, in any current or future media, including reprinting/republishing this material for advertising or promotional purposes, creating new collective works, for resale or redistribution to servers or lists, or reuse of any copyrighted component of this work in other works.}%
\thanks{This work was supported by KUKA Deutschland GmbH and conducted at KUKA premises in Augsburg, Germany.}%
\thanks{$^{1}$Technology \& Innovation Center, KUKA, Augsburg, Germany}%
\thanks{$^{2}$Institute of Robotics and Process Control, Technische Universität Braunschweig, Braunschweig, Germany}%
}

\begin{document}

\maketitle
\thispagestyle{empty}
\pagestyle{empty}

\begin{abstract}

Kinesthetic teaching through robot hand-guiding provides a natural interface 
for collecting demonstrations in imitation learning and 
programming-by-demonstration. However, extended sessions cause 
operator fatigue, reducing demonstration quality and limiting scalability. 
Current industrial hand-guiding approaches typically 
provide no active assistance, and alternatives require
costly wrist-mounted force-torque sensors or rely on learned motion priors 
unavailable for new tasks. We propose RHOAS, a hand-guiding scheme that 
actively supports operator-intended motions using model-based force 
estimation without additional hardware. 
Our approach considers robot hand-guiding as an actively controlled interaction by the human operator, rather than an interaction with a passive environment. Standard methods used for
hand-guiding typically rely on general passivity-based compliant
control architectures that unnecessarily increase operator effort and limit the range of demonstrable motions without providing the intended stability guarantees
in active interaction. Instead, our design utilizes model-based external torque
estimation, internal joint torque sensing, and redundant robot kinematics to actively
support human physical input within the human interaction frequency bandwidth.
We address 
practical challenges of relying on observer-based force estimation, including  suppression of unmodeled joint elastic dynamic effects and
measurement noise in the feedback path, reduced estimate accuracy close to kinematic singularities, and static gravity compensation errors. 
In a user study with 16 participants on a KUKA LWR iiwa we demonstrate
statistically significant reductions in physical effort,  
improved maneuverability for both precise and agile tasks, and clear user 
preference.

\end{abstract}

\section{INTRODUCTION}

\begin{figure*}[t]
    \centering
    \begin{subfigure}[t]{0.25\textwidth}
        \centering
        \includegraphics[width=\linewidth]{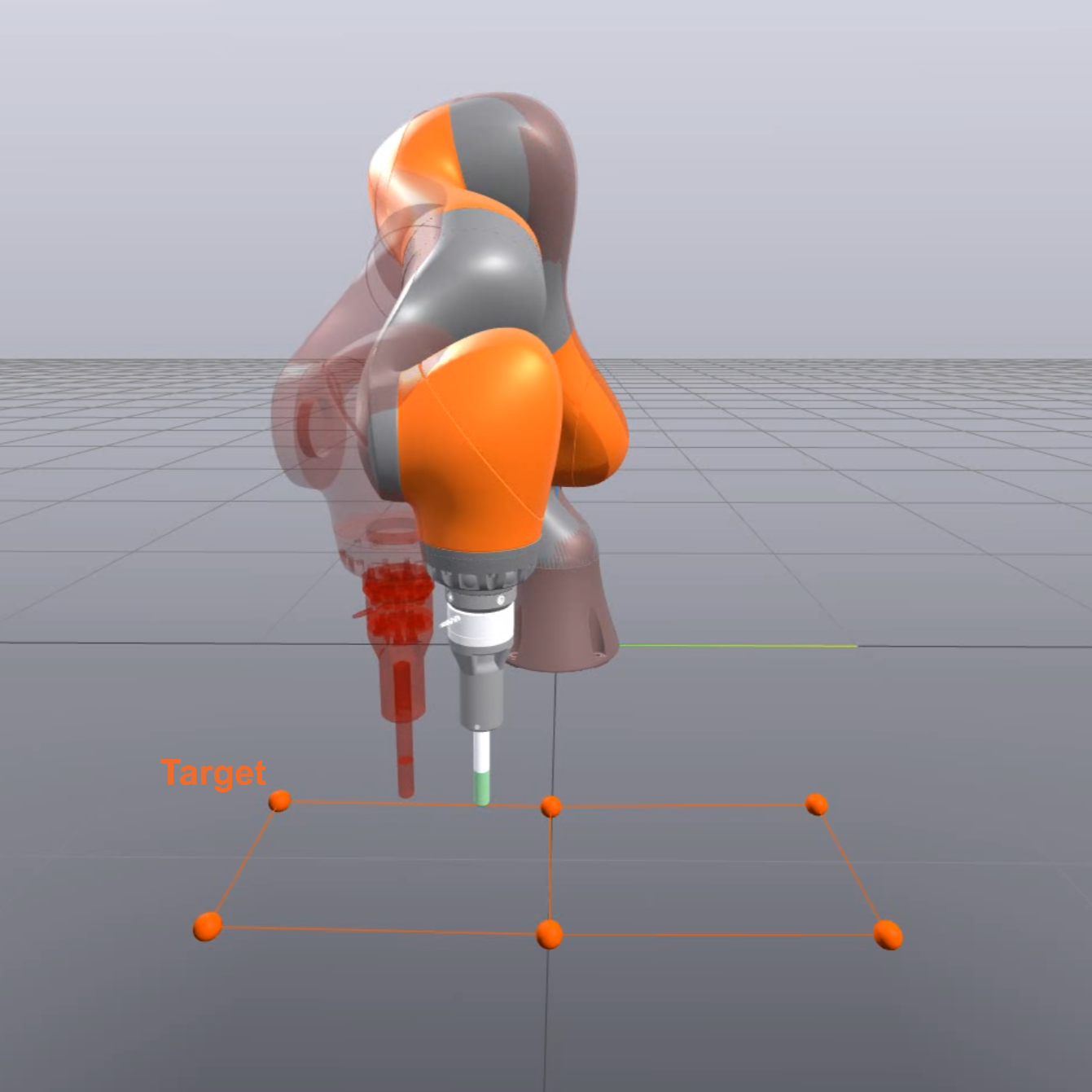}
        \caption{Estimate-based assisted\\ acceleration towards target}
    \end{subfigure}\hfill
    \begin{subfigure}[t]{0.25\textwidth}
        \centering
        \includegraphics[width=\linewidth]{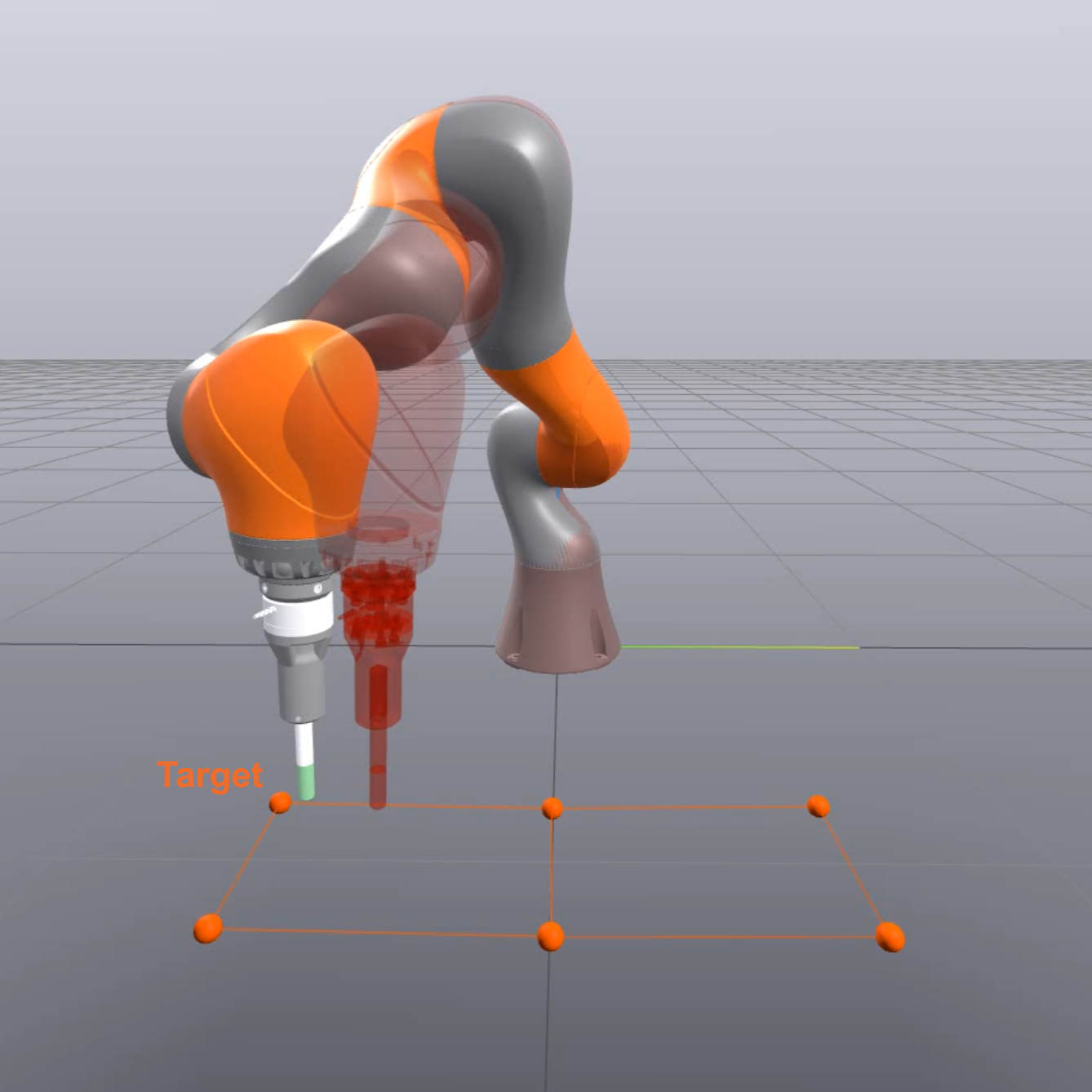}
        \caption{Assisted deceleration\\ before reaching target}
    \end{subfigure}\hfill
    \begin{subfigure}[t]{0.25\textwidth}
        \centering
        \includegraphics[width=\linewidth]{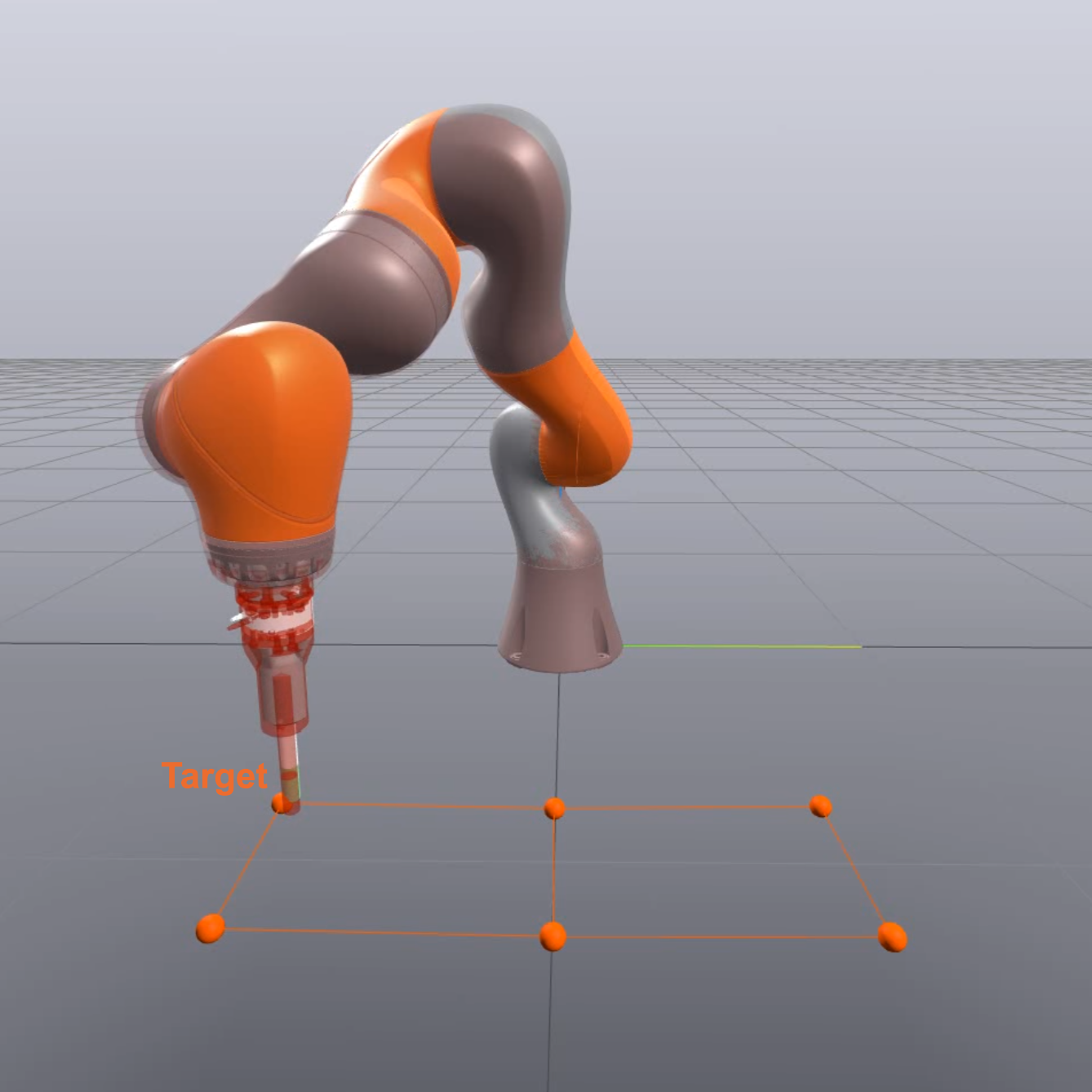}
        \caption{Converging at \\ target location}
    \end{subfigure}\hfill
    \begin{subfigure}[t]{0.25\textwidth}
        \centering
        \includegraphics[width=\linewidth]{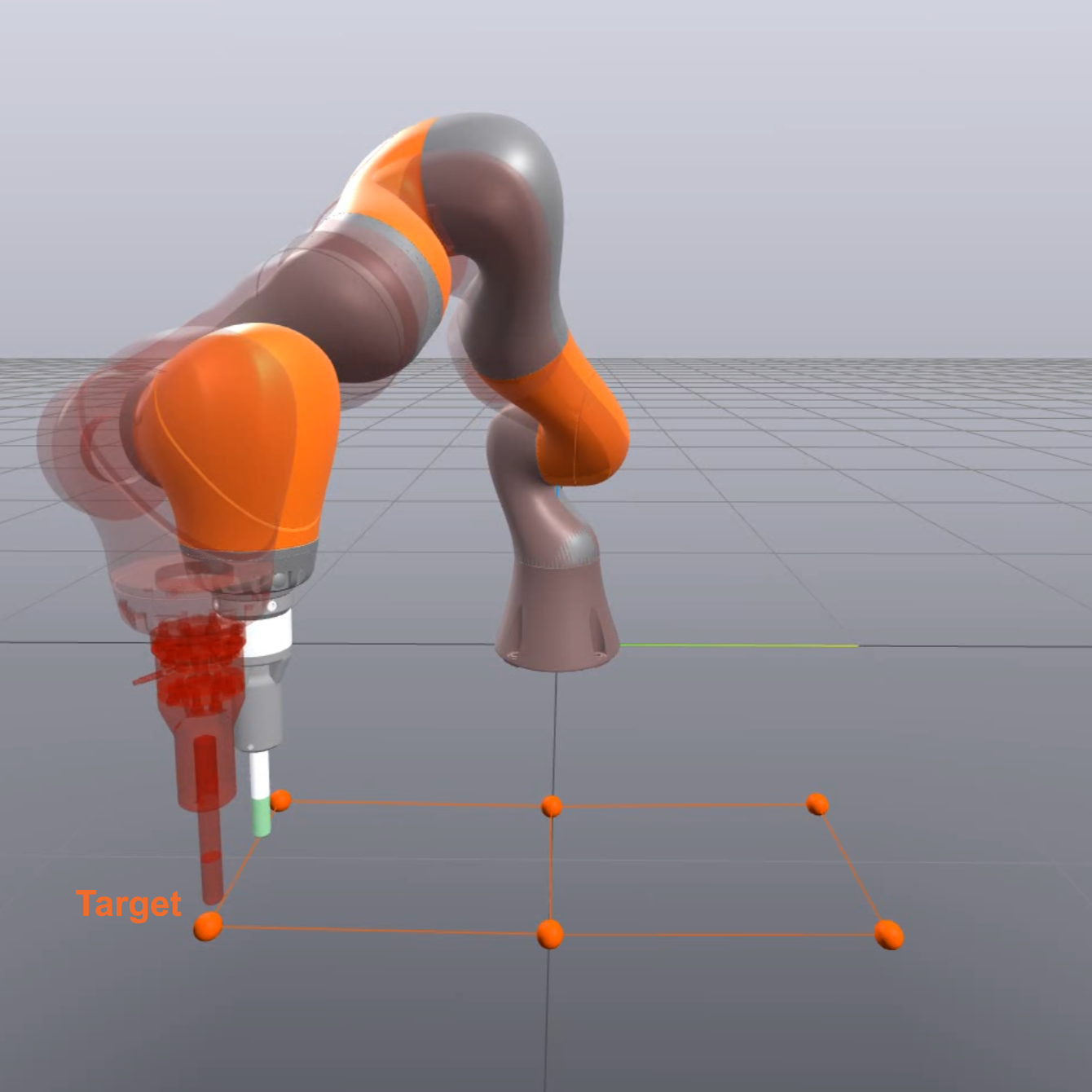}
        \caption{Assisted acceleration\\ towards next target}
    \end{subfigure}
    \caption{Visualization of the proposed active support during 
        a hand-guiding experiment (see supplementary video). 
        The transparent robot shadow shows the effect of the 
        supportive torque action by displaying a proxy of the velocity 
        setpoint~\eqref{eq:velctrl} mapped to robot joint space. The shadow 
        intuitively reveals how the observer-based processing 
        pipeline translates operator-applied forces into 
        responsive motion support — assisting both acceleration 
        and deceleration phases without noticeable artifacts 
        from modelling inaccuracies or estimation delays.}
    \label{fig:teaser}
\end{figure*}

Collaborative robots feature a hand-guiding mode that provides a natural interface for programming complex tasks through kinesthetic teaching~\cite{ravichandarRecentAdvancesRobot2020}. 
This mode is increasingly important as data-driven robot 
learning methods — including imitation
learning and programming-by-demonstration — frequently require large demonstration 
datasets, for which hand-guiding offers an intuitive and 
accessible approach.
In practice, however, collecting such data can take several hours of interaction, particularly in scenarios with high process or environment
variance~\cite{meli_robot_2025}. During extended sessions, operator fatigue becomes a critical limitation: it increases physical strain and progressively degrades the quality and consistency of demonstrations~\cite{peternel_fatique_2016}.

The standard industrial hand-guiding modes compensate for gravity effects, 
while (for some manipulators) an internal high-frequency feedback reduces the apparent inertia of the motors~\cite{passivityControlFlexRobots}. 
Consequently, operators can interact safely as the robot responds to external forces and moves accordingly.
However, gravity compensation provides no active assistance, leaving the operator to perform all the work of initiating and directing the motion against the robot's inertia, which becomes exhausting over time.
For redundant manipulators, the need to simultaneously manage TCP and nullspace motion further increases the difficulty~\cite{wredeUserStudyKinesthetic2013}. 

To address this challenge, several works have utilized wrist-mounted force-torque~(FT) sensors to infer motion intent and enhance the hand-guiding experience. Measured interaction forces have been employed in adaptive~\cite{gopinathanUserStudyPersonalized2017,hamadAdaptiveHumanForce2021}, 
learning-based~\cite{yacoubImprovingHumanRobot2021,nicolisHumanIntentionEstimation2018}, 
and
model-based~\cite{gregorHandGuidingVirtual2021,shaoConstraintAwareIntentEstimation2024}
and 
hybrid~\cite{haningerModelPredictiveControl2022,lawitzkyFeedbackMotionPlanning2012} 
controllers designed to alleviate operator effort. Despite decreasing FT sensor prices, these devices remain costly relative to collaborative robot platforms, making the high-quality kinesthetic teaching at scale particularly expensive.

External forces applied to the robot can also be estimated without direct measurement using the dynamic robot model and integrated joint torque sensors. 
This has been extensively studied for collision detection~\cite{haddadinRobotCollisionsSurvey2017}, with numerous observers~\cite{delucaCollisionDetectionSafe2006,garofaloSlidingModeMomentum2019} and estimators~\cite{wahrburgCartesianContactForce2015,manuelliLocalizingExternalContact2016} established in research and industry. 
For robots with integrated elastic joint torque sensors, 
motor-side friction is separated from the external torque in the link dynamics, 
making the estimate significantly more suitable for inferring user intentions~\cite{leeTorqueControlBased2016,zhangSensorlessHandGuiding2019}. 
Prior work on sensorless hand-guiding uses such estimates to enable basic guidance 
on robots lacking joint torque sensors by compensating friction~\cite{leeTorqueControlBased2016,zhangSensorlessHandGuiding2019}
— a different problem from actively supporting operator-intended motions. To our knowledge, observer-based force estimation has not been applied for closed-loop active motion support during hand-guiding.

Another group of methods, applicable for hand-guiding, relies on general compliant control architectures, where passivity-based design ensures stable behavior under the assumption that the environment is passive~\cite{passivityControlFlexRobots,ottCartesianImpedanceControl2008}. However, hand-guiding is a special case of compliant interaction in which the human operator actively controls the robot throughout the demonstration, rather than constituting an unknown passive environment.
Since the passivity assumption is not satisfied during hand-guiding, the associated conservatism unnecessarily increases operator effort and limits the range of demonstrable motions, without providing the intended stability guarantees.

We propose \mbox{RHOAS} (Robot Hand-guiding with Observer-based Active Support), 
a hand-guiding scheme built on three key design principles. 
First, we consider hand-guiding as a special case of compliant interaction where the interacting system is not passive, but actively controlling the robot.
Second, we design the controller to actively support the human actions in the range
of their motor capabilities, in terms of both response bandwidth and gain magnitude; this is achieved through a combination of response frequency filtering and a suitable control architecture. 
Third, we utilize the momentum-based force estimation and leverage it's advantages such as
lower noise sensitivity and suitability for robotic arms with direct joint torque sensing, to achieve reliable and effective motion assistance.

Our contributions are:
\begin{enumerate}
    \item \textbf{Robot response shaped to match human motor capabilities}: by treating the human operator as the controller of the human-robot system rather than an unknown environment, we design the robot response to be well-suited for typical human motor and proprioceptive capabilities.
    \item \textbf{Reduced operator effort}: enabling extended teaching sessions and large-scale data collection by significantly lowering the physical effort.
    \item \textbf{Enabling demonstration of fast and agile motions}: otherwise very difficult with current gravity-compensation, passivity-based, or learning-based compliant control approaches.
    \item \textbf{Scalability}: high-quality motion data-collection with collaborative manipulators, without the additional cost of wrist-mounted FT sensors.
    \item \textbf{Improved demonstration quality}: enhanced precision, maneuverability, and user perception.
\end{enumerate}

The design is validated in a user study with 16 participants, demonstrating statistically significant reductions in physical effort ($p < 10^{-4}$), improved user preference ($p < 10^{-5}$), and substantially enhanced maneuverability in agile tasks.

\section{PRELIMINARIES}
\label{sec:preliminaries}

Our approach considers a fixed-base lightweight torque-con\-trolled robot manipulator that is equipped with motor encoders for joint position measurements and
joint-embedded torque sensors measuring elastic torque.
In this section, we  (i)~provide a concise overview of the robot dynamics
and (ii)~revisit the state-of-the-art momentum observer used for external force estimation.

\subsection{ROBOT DYNAMICS}

\begin{figure}
    \centering
    \includegraphics[width=\linewidth]{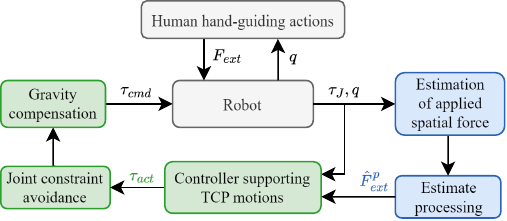}
    \caption{Overview of the proposed control scheme.
    To initiate desired robot motion,
    the operator applies forces that are estimated
    (Sec.~\ref{sec:externalTorqueObserver}) based on the measurement of joint position and joint elastic torque. This information is postprocessed~(Sec.~\ref{sec:processing}) and utilized to actively support the intended 
    movement  (Sec.~\ref{sec:control_architecture}).}
    \label{fig:high_level_system_overview}
\end{figure}

Consider a kinematic chain with~$n$~joints.
Link dynamics are modeled (in the reduced form~\cite{spongModelingControlElastic1987}) as 
\begin{align}
        \massMat \jointAcc + \ccMat \jointVel + \gravVec &= \jointTorque + \extTorque\,, \label{eq:linkElasticDynamics}
\end{align}
where
$\jointPos, \jointVel, \jointAcc \in \R^n$ are the joint positions, velocities and accelerations respectively,
$\massMat \in \R ^{n\times n}$ is the generalized inertia matrix, 
$\ccMat \in \R ^ {n \times n}$ is the matrix of Coriolis and centrifugal terms,
and~$\gravVec \in \R^n$ is the vector of torques due to gravity.
On the right-hand side, $\jointTorque \in \R^n$ is the joint elastic torque, containing the contribution of motor
torques and the effects of joint elasticity and motor dynamics. 
Lastly, $\extTorque \in \R^n$ is the external torque, resulting from the forces
applied to the robot by a human operator during hand-guiding, as shown in Fig.~\ref{fig:high_level_system_overview}, influenced by unmodeled dynamics and other disturbances.
In the following, we expect the forces applied by the human operator to be dominant to other mentioned effects,
and use their estimation to improve the hand-guiding interaction.
Eq.~\eqref{eq:linkElasticDynamics} assumes that joint damping and friction on the link side
of the transmission are negligible.

\subsection{Estimation of Externally Applied Spatial Force}
\label{sec:externalTorqueObserver}
To estimate the external torque~$\extTorque$ in~\eqref{eq:linkElasticDynamics}, we utilize 
the momentum observer approach
introduced by~\cite{delucaCollisionDetectionSafe2006} with the following form 
\begin{equation}
    \extTorqueEst = \observerGain \left[ 
    \momentumEst - \int_0^t
        \jointTorque + \extTorqueEst + \ccMatEst^T \jointVel - \gravityTorqueEst \
    dt
    \right]\,, \hspace{0.6cm} \label{eq:observer}
\end{equation}
where~$\momentumEst = \massMatEst \jointVel  - \bm{\hat{M}}(\bm{q_0})\bm{\dot{q}_0} \in \R ^n$ is the 
estimated generalized momentum.
This observer has the dynamics of $n$ decoupled first-order filters
and a time constant determined by a diagonal positive definite gain matrix $\observerGain \in \R^{n\times n}$.
In this formulation, we rely on the identified robot model in the form of the Coriolis and centrifugal
matrix~$\ccMatEst \in \R^{n \times n}$ and the gravity torque~$\gravityTorqueEst \in \R^n$,
and the availability of joint position~$\jointPos$, 
joint velocity~$\jointVel$ and joint elastic torque~$\jointTorque$ vectors. 

For the purpose of our hand-guiding control scheme, 
we focus on the case where the user is mainly applying forces at the  tool center point (TCP).
The spatial force~$\extForceEst \in \R^6$ applied at the tool is obtained from the estimated external torques~$\extTorqueEst$ through
\begin{align}
\label{eq:forceestimate}
    \extForceEst = \left(\bm{J}^T(\jointPos)\right)^{\#}\extTorqueEst\,,
\end{align}
using the Jacobian pseudoinverse solution~\cite{haddadinRobotCollisionsSurvey2017}.

It follows that the quality of the estimate ~$\extForceEst$ is significantly influenced by model inaccuracies, unmodeled dynamics, observer dynamics and other factors.
This fact makes it non-trivial to use~$\extForceEst$
in the context of physical human-robot interaction 
for hand-guiding with active support, and requires special care compared to common
FT sensor-based hand-guiding control architectures.

\section{APPROACH}
\label{sec:approach}

We introduce RHOAS, 
a control approach that actively supports operator-intended motions using 
estimated externally applied forces, without requiring a wrist-mounted FT sensor.
Our approach uses the estimated externally applied spatial force as feedback
to decrease the human effort required for moving the robot by hand.
We address the challenges involved in integrating the estimate in the control loop, particularly in relation to singular configurations and model inaccuracies (Sec.~\ref{sec:processing}).
Following this, we design a control scheme that leverages the estimate to actively assist the user's intended TCP motion~(Sec.~\ref{sec:control_architecture}).

\subsection{Control-Oriented Estimate Processing Pipeline}\label{sec:processing}

\subsubsection{Limiting the estimate and robot response bandwidth}\label{sec:filtering}
First, we limit the bandwidth of the force estimate~\eqref{eq:forceestimate} to human interaction frequencies and thereby avoid excitation of the unmodeled joint elastic
dynamic modes in the feedback loop.
According to the study by \cite{delice_intelligent_2007}, human control actions are generally restricted to frequencies no higher than $2$~Hz.
Therefore, we filter the estimated external force~$\extForceEst$ with a
low-pass filter at a cutoff frequency of 5 Hz to match the expected human bandwidth, yielding~$\extForceEstFiltered$. 
This sufficiently reduces the influence of high-frequency information that cannot be attributed to the human actions,
while avoiding unnecessarily increased resistance during typical human motions, and results in a more natural robot behavior for the operator. 

Note that low-pass filtering introduces delay, which could be problematic when
combined with a less efficient external torque estimation approach.
In our case, however, the total response delay remains below the threshold of human perception, ensuring that the operator's ability to control the robot and the perceived robot responsiveness
are not negatively affected.

\subsubsection{Confidence factor}\label{sec:confidence_factor}
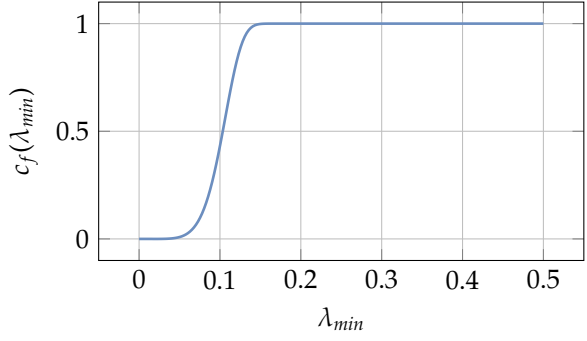
\begin{figure}[tbp]
    \centering
    \begin{tikzpicture}
        \begin{axis}[
            xlabel={$\singMin$},
            ylabel={$\confidenceFactor(\singMin)$},
            grid = both,
            domain=-0.0:0.5, 
            samples=200, 
            height=5cm,
            width=8cm,
        ]
        \addplot[color=graphBlue, line width=1pt]{1-exp(-(x/0.11)^6)};
        \end{axis}
    \end{tikzpicture}
    \caption{Confidence function for~$\confidenceThreshold = 0.11$, selected in the experiments.
    The function values close to zero~$\singMin$ scale down the estimate magnitude
    and thereby transition from the active support to pure zero-gravity mode.}
    \label{fig:confidence_function}
\end{figure}

Accounting for reduced accuracy of~$\extForceEst$,
near kinematic singularities is critical when integrating it within the control loop. Inspired by related works that modify singular values in projection operators~\cite{dietrichContinuousNullSpace2012}, we tackle this issue by introducing a configuration-dependent confidence factor~$\confidenceFactor(\singMin) \in \left [ 0, 1 \right ]$ to reduce the magnitude of estimated and filtered external spatial force~$\extForceEstFiltered$ as singular configurations are approached

\begin{align}
    \extForceEstScaled = \confidenceFactor(\singMin) \cdot \extForceEstFiltered \,.
\end{align}
This effectively disables the estimate-based motion support when the confidence is low, transitioning to pure zero-gravity torque mode\footnote{Gravity compensation is not affected by singularities.}. 
Our solution enables us to enhance the ease of hand-guiding
through active support in most regions of the workspace, while maintaining safe
behavior (e.g., gravity compensation only) near singularities.
The scalar confidence factor~$\confidenceFactor$ is determined via an exponential decay function applied to the smallest singular value~$\singMin \in \R$ of the TCP Jacobian matrix
\begin{align}
    \confidenceFactor(\singMin) = 1 - \exp{\left(-\left(\frac{\singMin}{\confidenceThreshold}\right)^6\right)} \,\,,
\end{align}
to ensure a smooth transition and sufficient flatness to suppress
the active control when~$\singMin$ exceeds the selected confidence
threshold~$\confidenceThreshold \in \R$.
Fig.~\ref{fig:confidence_function} shows the graph of the function for
a threshold of~$\confidenceThreshold = 0.11$.

\subsubsection{Estimate projection}\label{sec:projection}
Lastly, the estimate accuracy is impacted on low~frequencies
by the effect of gravity compensation imperfections\footnote{Imperfect compensation of Coriolis and centrifugal forces is negligible for slow velocities during handguiding.}, resulting in residual torque acting on the robot body.
This torque is regarded as external and, unless special precautions are taken, it is further magnified by the control architecture introduced in Sec.~\ref{sec:control_architecture}.
To mitigate amplification of this effect, we avoid using the external force estimate in the direction of the
gravity vector, achieved through an orthogonal projection
$\projectionXY=\text{diag}\{1, 1, 0, 0, 0, 1\}$ in the world-coordinate system %
\begin{align}
    \extForceEstProj &= 
    \projectionXY \, \extForceEstScaled\,,
\end{align}
selecting the estimate components parallel to the ground.
Ideally, the external force should be projected orthogonal to the anticipated gravity compensation error direction, taking into account the robot's dynamics. This approach may allow for further improvements in future work.

\subsection{Hand-guiding Control Architecture}\label{sec:control_architecture}

We here introduce a control scheme specifically designed
to utilize the postprocessed external spatial force estimate~$\extForceEstProj$ for actively supporting user motions. 
The aim of our architecture is to ensure that the hand-guiding is effortless for the user, while maintaining
intuitive and natural physical behavior.

\begin{figure}[tbp]
    \centering
    \includegraphics[width=\linewidth]{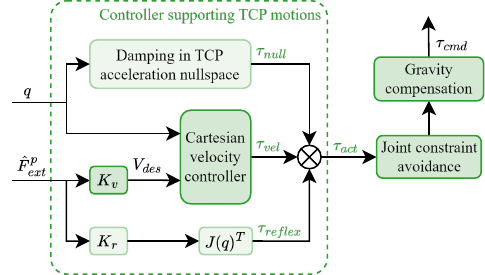}
    \caption{Proposed hand-guiding control architecture~\eqref{eq:final_command}.
    At the core is the Cartesian velocity controller (Sec.~\ref{sec:CV}), actively supporting TCP motion
    by controlling the spatial velocity to desired value, proportional to the estimated external forces. It can be extended with reflex action (Sec.~\ref{sec:reflex}), and with nullspace
    task for redundant robots (Sec.~\ref{sec:redundancy}). Additionally repulsive torques~\cite{osorioPhysicalHumanRobotInteraction2019} are added for avoiding
    joint constraints and gravity torques are compensated by the built-in low-level controller~\eqref{eq:rhoas}.}
    \label{fig:control_architecture}
\end{figure}

Fig.~\ref{fig:control_architecture} shows the proposed control structure, 
with the final command $\cmdTorque \in \R^n$ consisting of the actively 
supporting controller $\cmdActive \in \R^n$, composed of the velocity 
controller~$\CVvelcmd$ (Sec.~\ref{sec:CV}), reflex 
extension~$\CVRreflexcmd$ (Sec.~\ref{sec:reflex}), and nullspace 
damping~$\nullCmdProj$ (Sec.~\ref{sec:redundancy}):
\begin{align}
\label{eq:final_command}
    \cmdActive &= \CVvelcmd + \CVRreflexcmd +\nullCmdProj\,,
\end{align}
as well as the repulsive constraint avoidance torque $ \cmdTorqueRepulsive \in \R^n$ and gravity compensation torque~$\gravityTorqueEst$
\begin{align}
\label{eq:rhoas}
    \cmdTorque &= \cmdActive + \cmdTorqueRepulsive + \gravityTorqueEst\,.
\end{align}

\subsubsection{Cartesian Velocity Controller}\label{sec:CV}
The active support of the TCP motion is designed on the velocity level,
where the estimated and postprocessed external spatial force~$\extForceEstProj$ 
applied by the user is mapped to a desired TCP spatial velocity
$\admittanceVelSetpoint \in \R^{6}$ through a diagonal positive definite force-to-velocity gain matrix~$\CVfeedforwardGain \in \R^{6\times 6}$ as
\begin{equation}
    \admittanceVelSetpoint = \CVfeedforwardGain \, \extForceEstProj\,.
\end{equation}
A feedback controller maps the velocity error
through proportional gain matrix
$\CVfeedbackGain \in \R^{6\times 6}$ onto a reference acceleration,
and through spatial inertia matrix~$\spatialInertiaMat \in \R^{6\times 6}$ to obtain the
spatial force command. Then we obtain~$\CVvelcmd$ by converting the spatial force command to joint torques via TCP Jacobian
matrix transpose~$\J^T \in \R^{n\times6}$~as
\begin{equation}
\label{eq:velctrl}
    \CVvelcmd = \J^T \left[ \spatialInertiaMat \CVfeedbackGain (\admittanceVelSetpoint - \currentSpatialVel) \right] \,,
\end{equation}
where 
$
    \currentSpatialVel = \J \jointVel
$
is the current spatial velocity of TCP.
We deliberately avoid relying on the low-level position control interface that is commonly utilized for
admittance control implementation. When using the observer instead of a direct external force measurement,
the integration of the desired velocity for position-level control command leads
to the accumulation of model inaccuracies during estimation.
Furthermore, this integral action introduces an unnecessary delay, which combined with the observer delay, results in a negative impact on the overall response. 

Another advantage of our formulation is the unique representation of spatial
velocity, which avoids the need of choosing a particular orientation format.
In preliminary experiments, we confirmed that relying solely on velocity tracking
shows better performance in observer-based hand-guiding control compared to integration for position.

Furthermore, selecting a relatively slow response in \eqref{eq:velctrl}, which feels natural and safe for human operators, mitigates the stability issues related to force feedback in contact with a rigid environment.
As stated by~\cite{whitneyForceFeedbackControl1977a},
the closed-loop system stability with force feedback can be determined by the criterion
\begin{align}
    0 < T G K_e < 1\,, \label{eq:force_feedback_stability_crit}
\end{align}
where~$T$ is the sampling rate, $G$ is the force feedback gain, and~$K_e$ is the combined stiffness of the sensor and the
environment (in this case the human, any encountered objects and unmodelled dynamics or disturbances). 
The problematic characteristics of stiff
FT sensors~\cite{ottCartesianImpedanceControl2008} are mitigated by replacing
them with an observer,
limiting~$K_e$ even in contact with rigid objects. 
The combination of limited environment stiffness and low force feedback gain ensures sufficient robustness
of the hand-guiding controller in practical experiments, 
including collisions with rigid objects.

\subsubsection{Reflex extension}\label{sec:reflex}
A second component~$\CVRreflexcmd$ is dedicated to achieve active support via direct external force feedback. 
We denote this component as reflex due to its resemblance to the reflex reaction strategy in~\cite{haddadinCollisionDetectionReaction2008}.
This extension further increases the responsiveness of the robot to user actions and reduces the robot's resistance to movement during interaction.

The reflex action~$\CVRreflexcmd$ is computed based on the estimated externally
applied spatial force~$\extForceEst$, using
a diagonal reflex gain matrix~$\reflexGain \in \R^{6\times6}$ and the TCP Jacobian to map the action to the joint
torque command
\begin{align}
    \CVRreflexcmd = \J^T \reflexGain \, \extForceEst\,.
\end{align}

\subsection{Role of kinematic redundancy in control robustness}\label{sec:redundancy}
During the handguiding of redundant arms, the need for simultaneous handling of both tool and nullspace motion in a kinematically redundant subspace additionally increases handguiding difficulty \cite{wredeUserStudyKinesthetic2013}. 

In our design, we utilize this kinematic redundancy to our benefit by introducing an additional projected dissipative (damping) torque~$\nullCmdProj$ into the tool velocity nullspace.
This serves two purposes: firstly, we improve the controllability of the robot arm for the human who can still utilize full support of the observer in TCP velocity space, 
but does not need to fully control the rest of the motion. Secondly, this additional
dissipation improves overall control robustness against modeling inaccuracies or resonance modes
that could be excited through the observer-based feedback path.

\section{USER STUDY}
\label{sec:user_study}
The following user study was examined and approved in assessment process of KUKA Deutschland GmbH,
evaluating experiment methodology, safety and data protection. 
Consent from the human subjects in the research was obtained prior to each trial.
To ensure the safety of our subjects, we conducted a risk assessment,
implemented various safety measures and informed the subjects of any associated risks.

\subsection{Methodology and Demographics} 
The group of 16 user study participants was selected to the best of our ability to provide a wide
variety of physiological characteristics (height, strength),
gender representation, and various professional background
and experience with robotics.

The experiments (precision and agility) were performed in two
rounds with blind and randomly shuffled order of compared approaches for each user, in order to minimize any bias caused by execution sequence or naming.
In total, each user performed $32$ individual experiments,
resulting in a total of $256$ evaluated trials.

\subsection{Experiment Design}

Two experiments have been designed to assess the performance of our control scheme, each representing the opposite extremes of hand-guiding challenges. %

\begin{figure}[t]
    \centering
    \begin{minipage}{0.48\textwidth}
        \includegraphics[width=\textwidth]{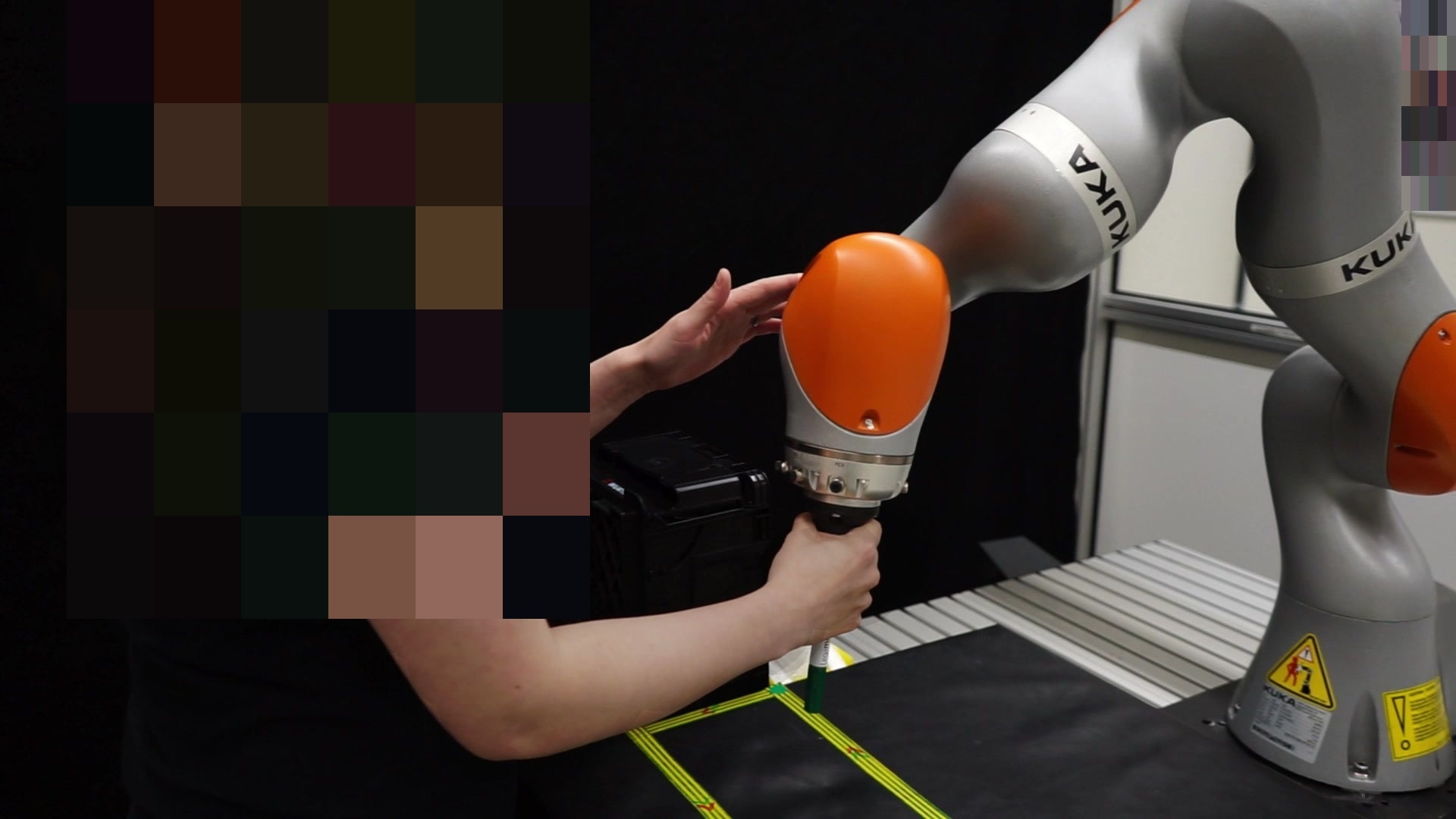}
    \end{minipage}\hfill%
    \begin{minipage}{0.48\textwidth}
        \includegraphics[width=\textwidth]{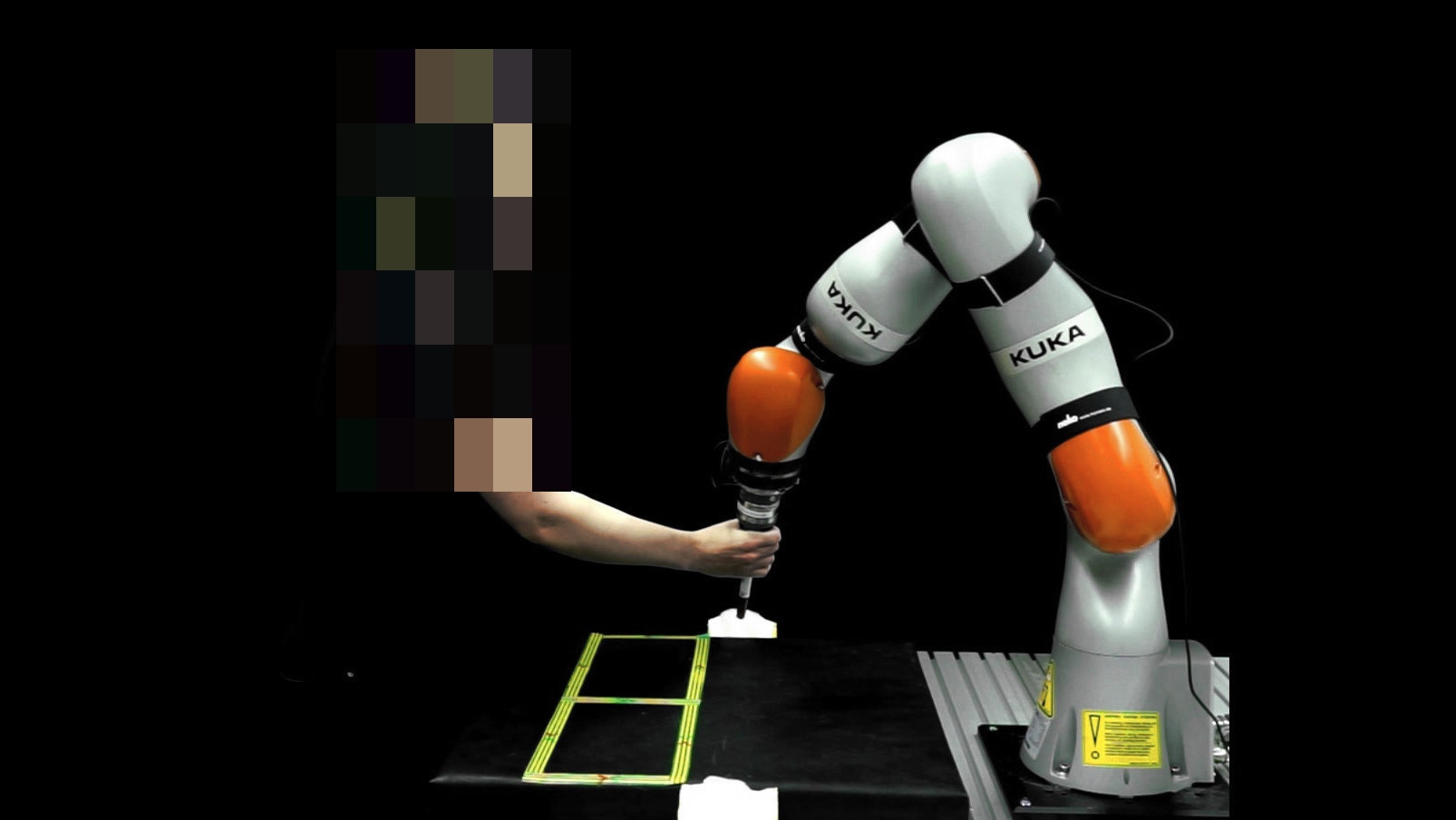}
    \end{minipage}
    \caption{Setup of precision and agility experiments. On the top is the precision experiment, where the user is guiding the robot to follow a reference path and adjusting the body with the other hand. On the bottom is the agility experiment, showcasing fast, repetitive movement with a pacer between target boxes.}
    \label{fig:experimental_setup}
\end{figure}

\textbf{Precision Experiment:} 
In the first experiment, we focus on precision-oriented motions, 
instructing users to trace a rectangular path using a marker mounted on our robot’s hand-guiding tool.
The path is outlined on a soft pad, depicted in Fig.~\ref{fig:experimental_setup}.
This experimental setup allows us to evaluate how our proposed method performs during both linear TCP motions and abrupt motion changes, which are challenging to execute accurately when hand-guiding articulated robots.

\textbf{Agility Experiment:}
In the second experiment, we investigate the behavior during larger, more agile motions.
Participants were asked to move the tool between two small boxes located approx. 60~cm apart from each other at a challenging pace of 45 bpm, set by a metronome, and contact with the attached pen the box on each beat (refer to Fig.~\ref{fig:experimental_setup}).
The challenging pace requires fast robot manoeuvrability and operates close to the limit of non-expert human motor capabilities.

\textbf{Experimental Platform:}
All experiments were conducted on the redundant 7~DoF KUKA LWR iiwa in the 7 kg payload version, controlled through Fast Research Interface (FRI)~\cite{schreiberFastResearchInterface2010} at $5$ms update rate, utilizing the gravity torque and friction compensation provided by the manufacturer.
To ensure a reliable metric calculation in the evaluation and verification of our control approach, irrespective of the quality of the external force estimation, we additionally measured the spatial force applied to the TCP using an ATI~mini~58 wrist-mounted FT sensor with NetBox for communication over Ethernet (not used within the control loop).
We further utilized the provided low-pass filter options and compensated the signal delay in postprocessing steps. 
We also designed a custom 3D-printed tool with an ergonomic grip
(see Fig.~\ref{fig:experimental_setup}).

\subsection{Compared Control Methods}
In the user study, we compare our RHOAS approach in two different variants
(w/wo reflex extension) with two baselines. In the following, we briefly introduce
the compared methods and their abbreviations. 

\begin{enumerate}
    \item \parbox[t]{\linewidth}{%
\textbf{CV} (Cartesian Velocity Controller):\\
Our observer-based approach as introduced in~\eqref{eq:rhoas} without the reflex action, i.e.~$\reflexGain=\bm{0}$,
supporting the TCP movement and suppressing motions in the nullspace with gains~$\CVfeedforwardGain = 0.15 \cdot \mathbf{I}_{6\times 6}$, and $\CVfeedbackGain = 0.7 \cdot \mathbf{I}_{6\times 6}$.}
\vspace{0.0\baselineskip}
    \item \parbox[t]{\linewidth}{%
\textbf{CVR} (Cartesian Velocity-Reflex Controller):\\
Extended version of the CV controller with identical gains~$\CVfeedforwardGain$ and~$\CVfeedbackGain$, including the reflex action in~\eqref{eq:rhoas} with~$\reflexGain  = 0.1 \cdot \mathbf{I}_{6\times 6}$.}
\vspace{0.0\baselineskip}
    \item \parbox[t]{\dimexpr\linewidth}{%
\textbf{G} (Gravity Compensation Controller):\\
Zero-gravity compensation scheme as provided by the low-level controllers of KUKA LWR iiwa with additional joint constraint avoidance on high-level, without any further high-level commands, i.e. ~$\CVfeedbackGain = \reflexGain = \bm{0}$ in~\eqref{eq:rhoas}.
Included as the standard approach to hand-guiding control.}
\vspace{0.0\baselineskip}
    \item \parbox[t]{\dimexpr\linewidth}{%
\textbf{CAD} (Cartesian Adaptive Damping Controller):\\
Alternative approach to our proposed controller design, inspired by~\cite{gopinathanUserStudyPersonalized2017}, however, based on the force estimate instead of a FT-sensor signal.
This controller utilizes the external spatial force in a passive way
for the adaptation of Cartesian damping. The adaptation is performed in
linear dependence on the magnitude of the force, decreasing the damping
when the applied forces are higher (to facilitate larger movements) and increasing
when the applied forces are lower (to make more precise movements easier to execute).}
\end{enumerate}

\subsection{Metrics}
We assess hand-guiding methods by analyzing three performance aspects:
human effort, user preferences, and task execution quality. 
To evaluate \textbf{human effort}, we compute the energy required to complete the task by integrating human power throughout the experiment execution: 
$\int_t \bf{f}(t)^T \, \bf{v}(t)$,
where~$\bf{v}(t)$ is the TCP twist,
and~$\bf{f}(t)$ is the external wrench measured by a wrist-mounted FT-sensor. 
In the precision experiment, this spans the entire reference path, while in the agility experiment, we evaluate movement between two boxes for 30 seconds at identical pace for all users.

Next, we evaluate the \textbf{user preferences} across the four control methods (CV, CVR, G, CAD). 
In each round of the experiment, the users provide a relative ranking of
controllers by ordering them from best~(1) to worst~(4) according to their preference
for the specific task.
To ensure unbiased ranking, the order of the controllers was unknown to the users during evaluation and randomized in each round.  

Lastly, we assess the \textbf{task execution quality}.
This involves evaluation of the mean distance deviation from the reference path for the precision experiment, and the mean time deviation from the reference pace on every metronome beat for the agility experiment.
Additionally, we track the number of beats that were missed completely during task execution due to user's difficulties with the desired pace.

\subsection{Experimental Results}
\begin{table}[t]
    \caption{Experimental results comparing the
    proposed (CV, CVR) and baseline (G, CAD) controllers across all evaluated metrics.
    Each value in table is reported as mean (standard deviation).
    Lower values indicate better performance for all metrics.
    Additionally, one-way ANOVA ($F$ statistic and $p$-value) is calculated for each metric based on all collected data.}
    \label{tab:descriptive}
    \centering
    \small
    \renewcommand{\arraystretch}{1.3}
    \begin{tabularx}{\columnwidth}{XXXXX}
        \hline
        & \textbf{CV} & \textbf{CVR} & \textbf{G} & \textbf{CAD}\\
        \hline
        \noalign{\smallskip}
        \multicolumn{5}{p{\dimexpr\columnwidth-2\tabcolsep\relax}}{\textit{Human effort}: Exerted energy (\%) normalized by the total energy exerted by individual user ($F$~=~$9.29$, $p$~$<$~$10^{-4}$).} \\[2pt]
        \textbf{Precision}
        & \textcolor{teal}{{24.2 (3.3)}} & \textcolor{teal}{{24.2 (5.7)}} & 25.4 (3.9) & 26.2 (5.2) \\
        \textbf{Agility}
         & 23.0 (3.7) & \textcolor{teal}{{22.8 (4.8)}} & 27.3 (3.5) & 26.9 (4.4) \\
        \textbf{Overall}
         & 23.6 (3.5) & \textcolor{teal}{{23.5 (5.3)}}  & 26.3 (3.8) & 26.5 (4.9) \\
         \hline
        \noalign{\smallskip}
        \multicolumn{5}{p{\dimexpr\columnwidth-2\tabcolsep\relax}}{\textit{Human effort}: Exerted energy (J) before normalization per user and experiment round ($F$~=~$0.45$, $p$~$>$~$0.05$).} \\[2pt]
        \textbf{Precision}
        & \textcolor{teal}{{20.8 (15.6)}} & 21.2 (18.0) & 21.1 (13.1)  & 23.3 (18.2) \\
        \textbf{Agility}
        & \textcolor{teal}{{183.5 (78.4)}} & 186.1 (94.7) & 222.0 (99.1)  & 209.0 (80.4) \\
        \hline
            
        \noalign{\smallskip}
        \multicolumn{5}{p{\dimexpr\columnwidth-2\tabcolsep\relax}}{\textit{User preference}: Relative controller ranking by users in each round ($F$~=~$9.71$, $p$~$<$~$10^{-5}$).} \\[2pt]
        \textbf{Precision}
          & \textcolor{teal}{{2.44 (1.03)}} & 2.59 (1.09) & 2.53 (1.09)  & \textcolor{teal}{{2.44 (1.20)}}\\
        \textbf{Agility}
         & 1.88 (0.79) & \textcolor{teal}{{1.66 (0.78)}} & 3.06 (0.97)  & 3.41 (0.79)\\
         \textbf{Overall}
         & 2.16 (0.96) & \textcolor{teal}{{2.13 (1.06)}} & 2.80 (1.09)  & 2.92 (1.12)\\
         \hline
         
        \noalign{\smallskip}
        \multicolumn{5}{p{\dimexpr\columnwidth-2\tabcolsep\relax}}{\textit{Task quality}: Count of missed metronome beats during single agility experiment ($F$~=~$0.51$, $p$~$>$~$0.05$).} \\[2pt]
        \textbf{Agility}
        & \textcolor{teal}{{1.16 (1.91)}} & \textcolor{teal}{{1.16 (2.55)}} & 1.56 (2.89) & 2.06 (4.16) \\
        \hline
    \end{tabularx}
\end{table}

\begin{table}[t]
    \caption{One-tailed Welch's t-test results comparing means of proposed and baseline controllers for statistically significant ANOVA results ($H_0: \mu_\textit{proposed} \geq \mu_\textit{baseline}$).}
    \label{tab:inferential}
    \centering
    \small
    \renewcommand{\arraystretch}{1.3}
    \begin{tabularx}{\columnwidth}{XXXX}
        \cline{1-4}
        \textbf{Proposed} & \textbf{Baseline} & \textbf{t-statistic} & \textbf{p-value}\\
        \hline
        \noalign{\smallskip}
        \multicolumn{4}{p{\dimexpr\columnwidth-2\tabcolsep\relax}}{\textit{Human effort}: Normalized exerted energy by human.} \\[2pt]
         CV & G & $-4.11$ & $3.53 \cdot 10^{-5}$ \\
         CV & CAD & $-3.81$ & $1.11 \cdot 10^{-4}$ \\
         CVR & G & $-3.42$ & $4.32 \cdot 10^{-4}$ \\
         CVR & CAD & $-3.31$ & $5.96 \cdot 10^{-4}$ \\
        \hline
        \noalign{\smallskip}
        \multicolumn{4}{p{\dimexpr\columnwidth-2\tabcolsep\relax}}{\textit{User preference}: Relative controller ranking by users.} \\[2pt]
         CV & G & $-3.50$ & $3.20 \cdot 10^{-4}$ \\
         CV & CAD & $-4.12$ & $3.41 \cdot 10^{-5}$ \\
         CVR & G & $-3.51$ & $3.07 \cdot 10^{-4}$ \\
         CVR & CAD & $-4.11$ & $3.55 \cdot 10^{-5}$ \\
        \hline
    \end{tabularx}
\end{table}\vfill%
In the following, we present both descriptive and inferential analysis of the experimental results.
The comparison of controllers in terms of mean and standard deviation of the observed metrics is reported in Tab.~\ref{tab:descriptive}.
Additionally, we performed a one-way Welch's ANOVA (analysis of variance) to assess the statistical significance of the results.
For the results with high significance based on ANOVA ($p < 0.05$) we further perform a pairwise one-tailed Welch's t-test (see Tab.~\ref{tab:inferential})
to assess individual controller's performance drop or increase against the baselines.

To account for differences in movement efficiency between individual users,
we normalize the exerted energy by the total exerted energy by the concrete user in the evaluated experiment round. 
With this, we can mitigate the variance in data caused
by individual user's capabilities and better isolate the impact of using the evaluated controllers.

\begin{figure}[tbp]
    \centering
    \includegraphics[width=\linewidth]{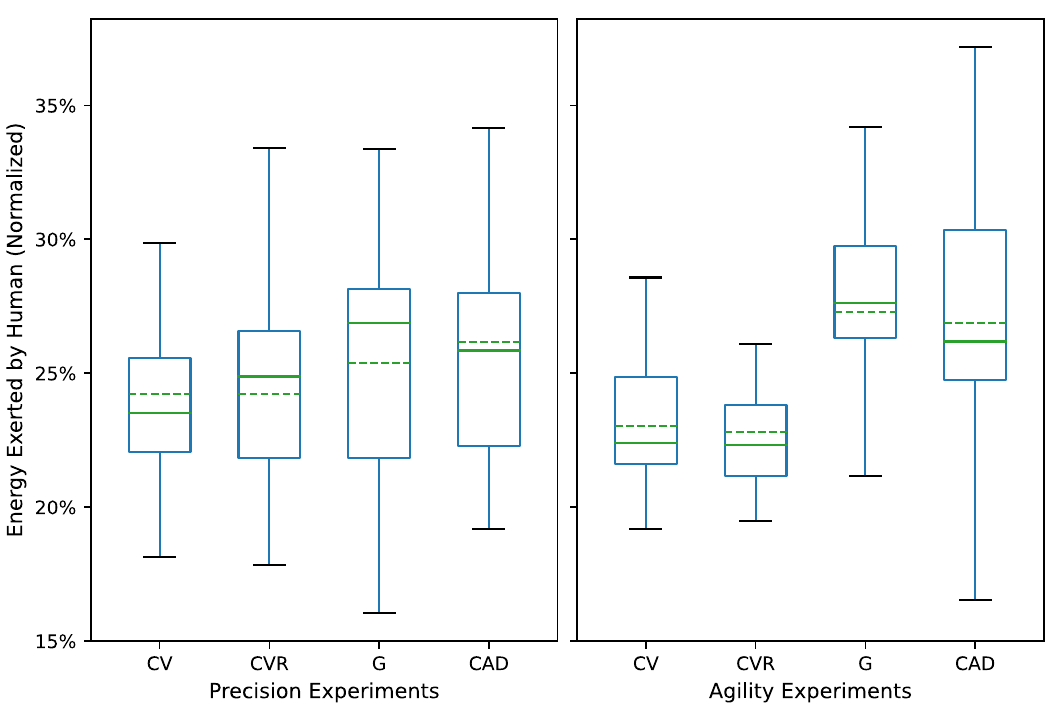}
    \caption{Distribution of normalized exerted human energy values.
    Both the CV and CVR controllers show improvement over the baselines, especially in the agility
    experiment.
    Data distribution is depicted with the quartiles and data range as
    solid lines and the mean value as green dashed line.}
\label{fig:normalized_energy_comparison_all_users_all_rounds}
\end{figure}

Both proposed controllers outperform the baselines in terms of reducing the required human effort.
The energy required to follow the trajectory from start to end, or to move the TCP between boxes at a specified pace, using CV and CVR controllers,
was lower than with pure gravity compensation (G)
and Cartesian adaptive damping controller (CAD) (see.~Tab.~\ref{tab:descriptive}).
This improvement suggests both positive impact on lowering the resistance to motion, as well as, better
controllability of the TCP movement. Merely relying on low motion resistance without reshaping the
robot's reaction in user's intended direction leads to more frequent and aggressive
path corrections. 
This effect was prevalent with the gravity compensation 
controller, where the user has to move a system with 
significant inertia unassisted, making it difficult 
to control the motion direction.
The distribution of normalized energy values from both experiments is shown in~Fig.~\ref{fig:normalized_energy_comparison_all_users_all_rounds}.

Also in terms of user preferences, the proposed controllers obtained a better mean rating
compared to the baseline methods, with a
high level of statistical significance ($p < 10^{-3}$) compared to the baselines as shown in Tab.~\ref{tab:inferential}. Both CV and CVR obtained comparable overall user 
preference, with CV favored in precision tasks and 
CVR in agility tasks.

Finally, the results also suggest improvement in terms of task execution quality, however, this trend could not be confirmed with statistical
significance based on the collected data.
The reduced number of missed beats while performing agility task at designated pace
would again correspond to
increased maneuverability obtained with CV and CVR controllers.
To validate this hypothesis and draw conclusions about other quality metrics,
a more extensive user study would be required to isolate
these effects from the individual differences in human performance.

\section{CONCLUSIONS}

\label{sec:conclusion}

We presented RHOAS, a hand-guiding control scheme that actively 
supports operator-intended motions using observer-based force 
estimation. Our design exploits the fundamental difference between 
hand-guiding and general compliant interaction: the human operator 
actively controls the robot, and the robot's response is shaped to 
match their motor capabilities. A user study with 16 participants 
confirmed statistically significant reductions in physical effort, 
improved maneuverability and clear user preference over baseline 
methods. By not requiring external sensors, RHOAS makes enhanced 
hand-guiding accessible for scalable kinesthetic teaching and 
large-scale demonstration collection. Desirable future work would 
include validation on other robot platforms, as well as an 
ablation study of the individual design components and parameters.

\bibliographystyle{IEEEtran}
\bibliography{references}

\end{document}